\documentclass[conference]{IEEEtran}
\IEEEoverridecommandlockouts

\usepackage{cite}
\usepackage{amsmath,amssymb,amsfonts}
\usepackage{graphicx}
\usepackage{textcomp}
\usepackage{xcolor}

\usepackage{siunitx}
\usepackage[caption=false,font=footnotesize]{subfig}
\usepackage{placeins}
\usepackage{multirow}
\usepackage[hidelinks]{hyperref}

\def\BibTeX{{\rm B\kern-.05em{\sc i\kern-.025em b}\kern-.08em
    T\kern-.1667em\lower.7ex\hbox{E}\kern-.125emX}}
\begin{document}

\title{A Feature Extraction Pipeline for Enhancing Lightweight Neural Networks in sEMG-based Joint Torque Estimation}

\author{\IEEEauthorblockN{Kartik Chari\IEEEauthorrefmark{1}\IEEEauthorrefmark{2}, Raid Dokhan\IEEEauthorrefmark{2}, Anas Homsi\IEEEauthorrefmark{2}, Niklas Kueper\IEEEauthorrefmark{1}\IEEEauthorrefmark{2}, Elsa Andrea Kirchner\IEEEauthorrefmark{1}\IEEEauthorrefmark{2}}
\IEEEauthorblockA{\IEEEauthorrefmark{1}\textit{Robotics Innovation Center},
\textit{German Research Center for Artificial Intelligence},
Bremen, Germany.}
\IEEEauthorblockA{\IEEEauthorrefmark{2}\textit{Institute of Medical Technology Systems},
\textit{University of Duisburg-Essen},
Duisburg, Germany.}
\IEEEauthorblockA{Email: Kartik.Chari@dfki.de}
\thanks{This work was funded by the German Federal Ministry of Education and Research (BMBF) under Grant No. 01IW24008.}
}

\maketitle

\begin{abstract}
Robot-assisted rehabilitation offers an effective approach, wherein exoskeletons adapt to users' needs and provide personalized assistance. However, to deliver such assistance, accurate prediction of the user's joint torques is essential. In this work, we propose a feature extraction pipeline using 8-channel surface electromyography (sEMG) signals to predict elbow and shoulder joint torques. For preliminary evaluation, this pipeline was integrated into two neural network models: the Multilayer Perceptron (MLP) and the Temporal Convolutional Network (TCN). Data were collected from a single subject performing elbow and shoulder movements under three load conditions (0 kg, 1.10 kg, and 1.85 kg) using three motion-capture cameras. Reference torques were estimated from center-of-mass kinematics under the assumption of static equilibrium. Our offline analyses showed that, with our feature extraction pipeline, MLP model achieved mean RMSE of 0.963 N m, 1.403 N m, and 1.434 N m (over five seeds) for elbow, front-shoulder, and side-shoulder joints, respectively, which were comparable to the TCN performance. These results demonstrate that the proposed feature extraction pipeline enables a simple MLP to achieve performance comparable to that of a network designed explicitly for temporal dependencies. This finding is particularly relevant for applications with limited training data, a common scenario patient care.
\end{abstract}


\section{Introduction}
True human-robot collaboration requires co-adaptation, where both humans and robots adapt to each other's behaviors. Traditionally, assistive robots contributed with physical strength while humans provided cognitive skills \cite{rahman2006passive, lovrenovic2019development, huang2020joint}. Today, robots are increasingly expected to simulate human behaviors, recognize actions, and anticipate intentions \cite{ji2019survey}, but human variability and physiological unpredictability remain major challenges.

In rehabilitation, accurate tracking and prediction of joint movements are critical. Some studies measured joint torques directly using sensors and mechanical devices \cite{hwang2015method}, while others relied on musculoskeletal models such as the Hill-type model \cite{lloyd2003emg, holmes2006teaching}. However, as noted in \cite{huang2020joint, naeem2012emg}, such models are difficult to construct, since they require numerous parameters to be measured using specialized equipment. Surface electromyography (sEMG)-driven approaches simplify computation by combining muscle models with body dynamics \cite{shao2009electromyography}. Building on this, numerical methods were developed to associate predefined hand motions with corresponding forearm EMG signals \cite{lisi2011classification}. 

However, these methods struggled to capture nonlinear muscle dynamics. As a result, machine learning (ML) approaches gained attention for joint torque estimation. For instance, \cite{naeem2012emg} used a backpropagation neural network (BPNN) to predict elbow torques from EMG whereas \cite{9257478} estimated ankle torques using an EMG-driven neuromusculoskeletal (NMS) model combined with an artificial neural network (ANN). More recent works applied recurrent convolutional networks (RCNN) to predict knee torques \cite{schulte2022multi, huang2019real}, while EMG-driven approaches with NMS models have been explored for the upper limb but they require precise parameter estimation, limiting their practicality \cite{tahmid2023upper}.

To address this, EMG-based feature extraction methods have emerged. Some studies have used statistical methods and manually extracted features to capture muscle dynamics and improve signal interpretation \cite{kok2024machine, phinyomark2014feature}. Autoencoders have been used to extract informative features \cite{huang2020joint}, convolutional neural networks (CNN) and temporal convolutional networks (TCN) have demonstrated superior performance in modeling temporal dependencies without explicit feature engineering \cite{chen2022real, lea2017temporal}. Nevertheless, these methods often require large datasets, complex pipelines, and computationally expensive architectures, which are impractical in rehabilitation contexts due to patient fatigue and limited data.

Motivated by these challenges, we propose a dedicated EMG feature extraction pipeline for estimating joint torques. Our pipeline enabled accurate predictions even with a simple multilayer perceptron (MLP) network. To evaluate whether the extracted features effectively capture temporal dynamics, we compared their performance with that of TCN, a model explicitly designed for sequence modeling using raw, preprocessed time points \cite{zanghieri2021semg}.

\section{EXPERIMENTAL DESIGN} \label{sec:expt_design}
This section provides details about the participant, the experimental setup, data acquisition methods, and the experimental procedure.
    \subsection{Participant} \label{subsec:participant}
    One healthy male participant (24 years old, student) voluntarily took part in this study. A few days before the experiment, he was familiarized with the setup and procedure. He provided written informed consent, including permission for recording and use of his images and videos. The participant was informed of his right to stop the experiment at any time without consequences. The study was approved by the local ethics committee and conducted in accordance with the local legislation and institutional requirements.

    \subsection{Experimental Setup} \label{subsec:expt_setup}
\begin{figure}[htbp]
        \centering
        \parbox{0.49\columnwidth}{
            \subfloat[Participant performing a task\label{fig:expt_setup_a}]{\includegraphics[width=\linewidth, trim={0 91mm 0 10mm}, clip]{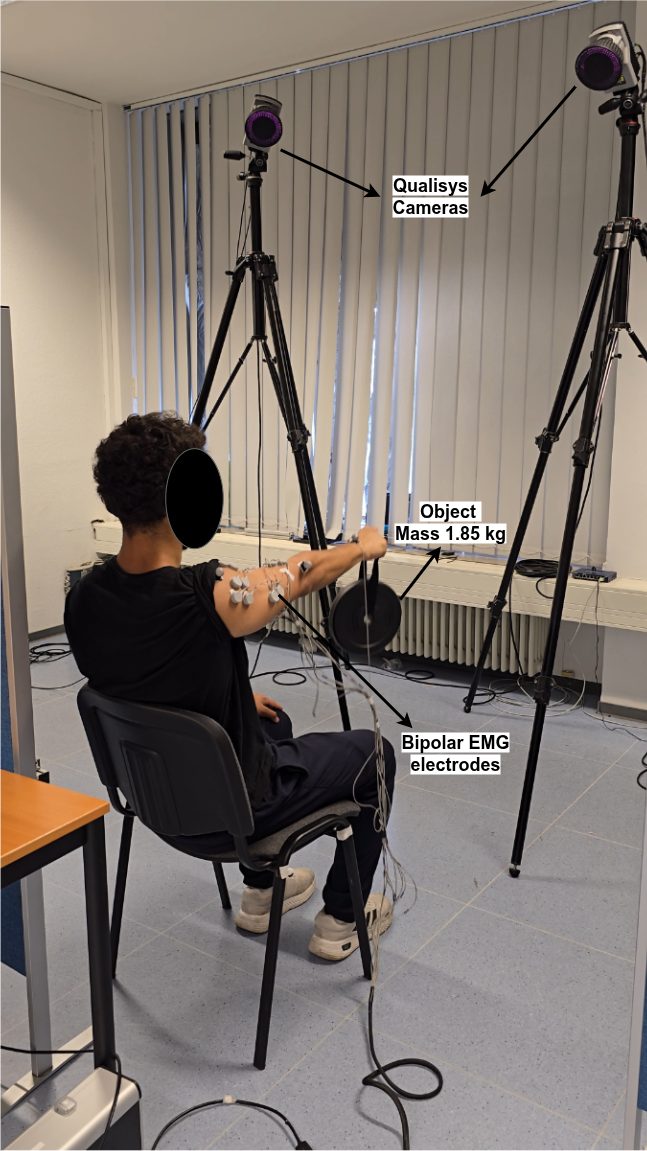}
            }
        }
        \hfill
        \parbox{0.49\columnwidth}{
            \subfloat[Objects used as weights\label{fig:expt_setup_b}]{\includegraphics[width=\linewidth, trim={1mm 0 0 0}, clip]{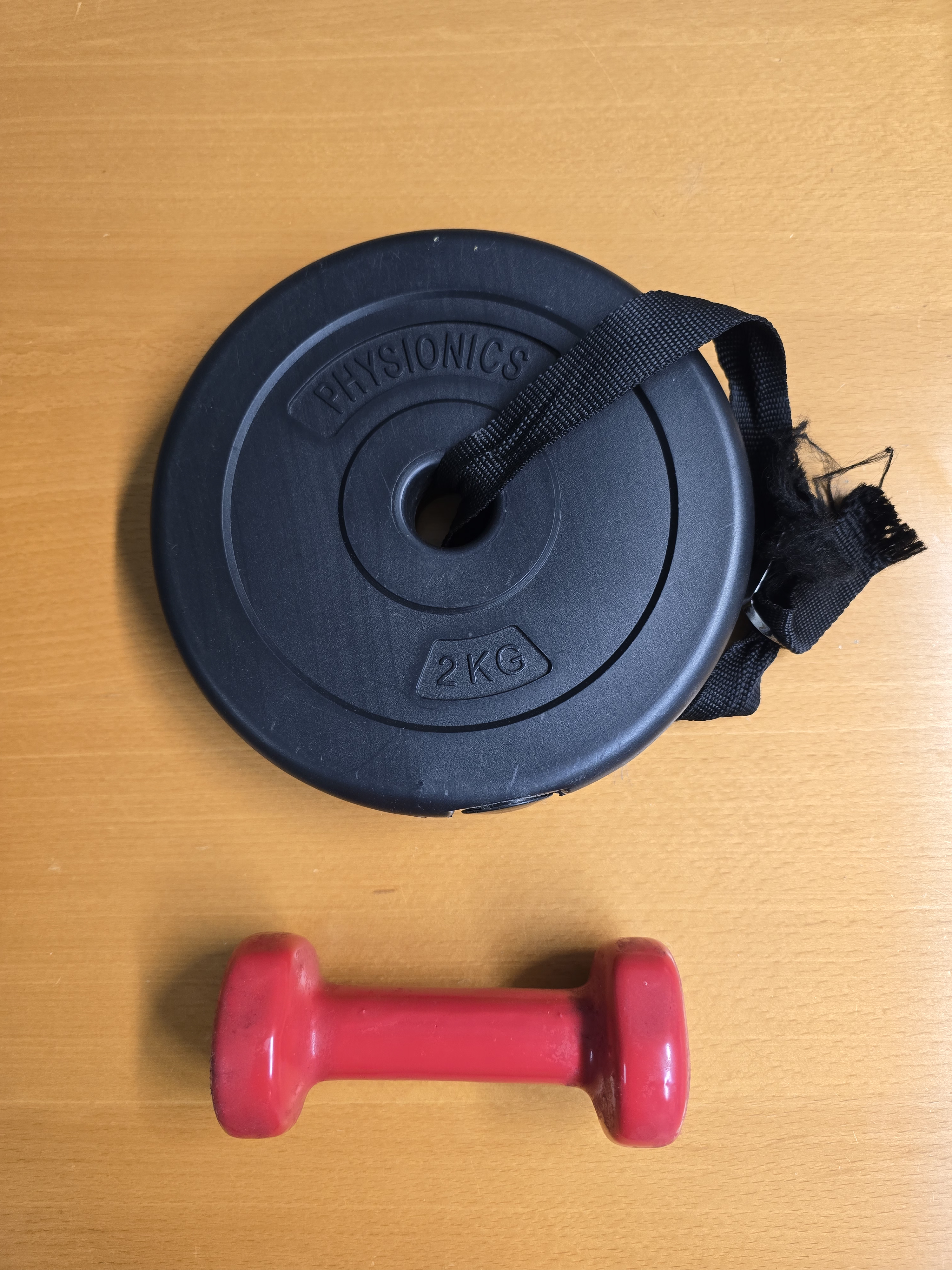}
            }
        }
        \caption{Experimental setup for data collection. (a) Participant prepared with EMG electrodes lifting an object; two of the three Qualisys motion-capture cameras are visible. (b) Weighted objects used in the experiment with measured masses of \SI{1.10}{\si{kg}} and \SI{1.85}{\si{kg}}. The \SI{0}{\si{kg}} weight is not shown.}
        \label{fig:expt_setup}
    \end{figure}

    Figure~\ref{fig:expt_setup_a} shows the setup used for data collection. The participant sat on a comfortable chair holding an object of mass \SI{0}{\si{kg}}, \SI{1.10}{\si{kg}}, or \SI{1.85}{\si{kg}}. The objects used as weights in the experiment are shown in Figure~\ref{fig:expt_setup_b}. Three Qualisys Oqus 300 cameras tracked the participant's joint trajectories with the help of six passive markers at a sampling rate of \SI{500}{\si{Hz}}. These markers were placed on the shoulders, elbows, and wrists of both arms, and the cameras tracked them from multiple viewpoints to obtain precise 3D positions of each marker. 
    Furthermore, the ANT mini eego amplifier was used to record bipolar surface EMG data at a sampling rate of \SI{500}{\si{Hz}} with the help of eego SDK for Python. 
    Eight EMG channels were recorded, and the electrodes were placed on the belly of each muscle after preparing the skin with 70\%~\text{v/v} isopropyl alcohol, in accordance with the SENIAM guidelines \cite{hermens2000development}. The electrodes measured the following muscles of the right arm: deltoid pars clavicularis, deltoid pars acromialis, deltoid pars spinalis, biceps brachii, triceps brachii caput laterale, triceps brachii caput mediale, brachialis, and pectoralis major.
    
    \subsection{Experimental Procedure} \label{subsec:expt_procedure}
    \begin{figure}[!htbp]
      \centering
      \includegraphics[width=0.8\columnwidth, trim={5mm 8mm 5mm 8mm}, clip]{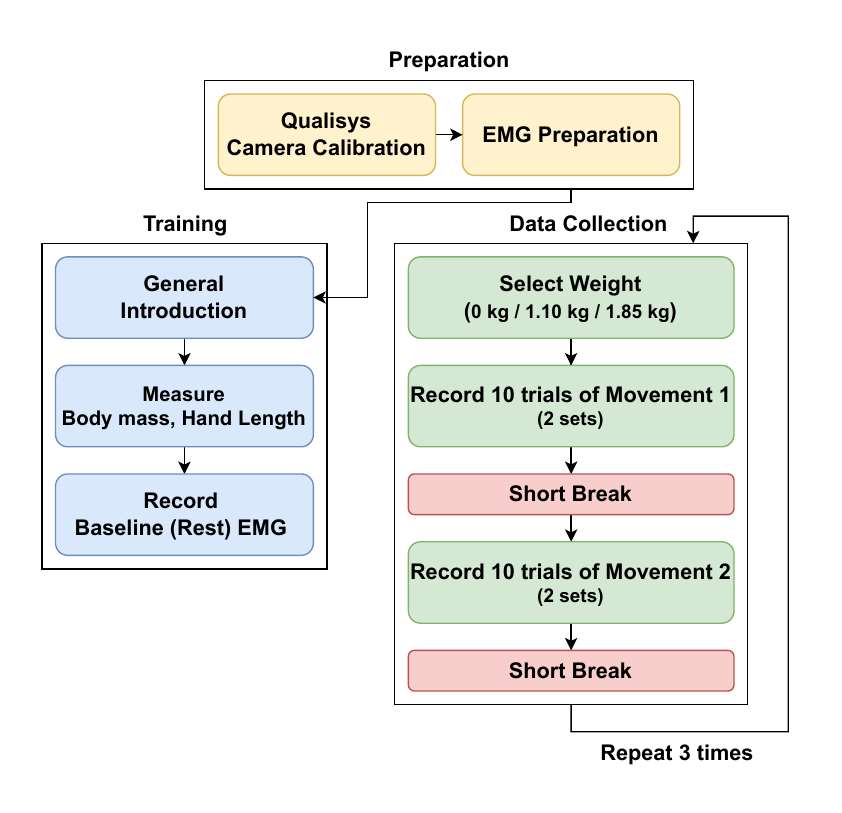}
      \caption{Flowchart of the experimental procedure. In each repetition, the participant performed the movements with weights chosen in ascending order. Movement 1 denotes grasping, and Movement 2 denotes complex movement.}
      \label{fig:expt_procedure}
    \end{figure}
    The aim of this experiment was to collect EMG and motion data from a healthy participant under varying levels of muscle activation in the shoulder and elbow. These data were intended as inputs to a machine learning regressor model for predicting human joint torques from their EMG signals, with motion data providing the reference torque values for supervised learning. To elicit these varying muscle activation levels, we selected three different weights (\SI{0}{\si{kg}}, \SI{1.10}{\si{kg}}, and \SI{1.85}{\si{kg}}) and two movement types, namely, grasping and complex movement. 
    
    The grasping movement consisted of moving the arm from rest to full extension in the sagittal plane and back, mimicking the action of grasping an object on a table. Similarly, the complex movement combined elbow flexion/extension with shoulder flexion/extension, as well as abduction and adduction, thereby capturing movements of the relevant degrees of freedom for our joint torque prediction use case. Figure~\ref{fig:expt_procedure} shows the procedure used to collect the multimodal data for offline analysis and model training, which was organized into three phases: preparation, training, and data collection.
    
     In the preparation phase, the Qualisys cameras were first calibrated to fixate their positions relative to the chosen frame of reference, ensuring accurate tracking within the calibrated workspace. Then, eight channels of bipolar surface EMG electrodes were placed on the participant's right arm. During the training phase, the experimental procedure was clearly explained to the participant, followed by the measurement of his body mass and hand length (distance between the wrist and the third finger's metacarpophalangeal joint (MCP)). The participant was then seated on the chair as shown in Figure~\ref{fig:expt_setup_a}, and baseline resting EMG was recorded. This baseline data was later used to correct offsets and drifts in the EMG signals. The data collection phase consisted of three sections. In each section, the participant held one of the weights (\SI{0}{\si{kg}}, \SI{1.10}{\si{kg}}, or \SI{1.85}{\si{kg}}) in ascending order and performed two sets of 10 grasping movement trials, followed by a short two-minute break, and then two sets of 10 complex movement trials. After each set, the Qualisys and EMG data were saved, and sample counts were verified to ensure complete recordings. The entire experimental session lasted approximately 40 minutes.

\section{METHODS} \label{sec:methods}
This section presents the methods used in this work, including the computation of reference joint torques, the design of the proposed feature extraction pipeline, and the machine learning models used to evaluate predictive performance.

    \subsection{Reference Joint Torques Computation}
    The reference joint torques served as targets for supervised learning. In this study, three joint torques were considered: elbow torque ($\tau_e$), and the projections of the shoulder torque onto the sagittal and frontal planes, named as front-shoulder ($\tau_{sf}$) and side-shoulder ($\tau_{ss}$) torques. These torques were computed from the measured arm kinematics and the gravitational forces using the static equilibrium method \cite{just2017feedforward}. Each arm segment was modeled as a rigid body with mass concentrated at its center of mass (COM). At every time instant, the system was assumed to be in static equilibrium, neglecting its dynamics. The wrist rotation was also considered negligible. 
    
    \begin{figure}[tb]
      \centering
      \includegraphics[width=0.75\columnwidth]{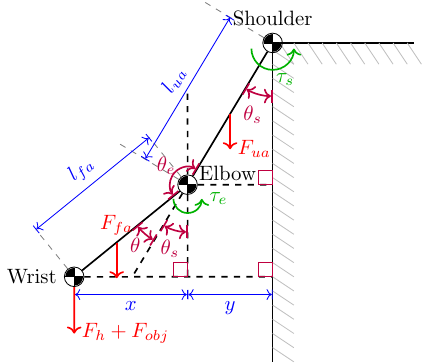}
      \caption{FBD of right arm. Hand is omitted for clarity.}
      \label{fig:fbd}
    \end{figure}
    Figure~\ref{fig:fbd} shows the free-body diagram (FBD) of the right arm. The perpendicular distances were calculated as 
    \begin{math}
            y = l_u \cdot sin(\theta_s), \ 
            x = l_{fa} \cdot sin(\theta_e - \theta_s)
    \end{math},
    where $y$ is the perpendicular distance between the shoulder and elbow, and $x$ is the perpendicular distance between the elbow and wrist. The angles $\theta_s$ (shoulder adduction angle) and $\theta_e$ (elbow extension angle) were computed from the Qualisys kinematic data using vectors connecting the shoulder, elbow, and wrist, as well as a vector connecting the right and left shoulders. Using the kinematic data along with segment masses and COM distances (see Tables~\ref{tab:segment_masses} and~\ref{tab:com_distance}), the participant's physiological parameters were derived.
    \begin{table}[htbp]
        \caption{Estimated arm segment masses as a percentage of total body mass for male and female individuals\cite{seg_weights_com}.}
        \label{tab:segment_masses}
        \centering
        \begin{tabular}{|l|c|c|}
            \hline
            \shortstack{Arm segment \\ (name)}& \shortstack{\\Male \\ (\% body mass)}& \shortstack{Female \\ (\% body mass)}\\
            \hline
            Upper arm& 2.71& 2.55\\
            Forearm& 1.62& 1.38\\
            Hand& 0.61& 0.56\\
            \hline
        \end{tabular}
    \end{table}

    \begin{table}[htbp]
        \caption{Estimated distance of the COM of each arm segment from its proximal joint origin, expressed as a percentage of segment length for male and female individuals \cite{seg_weights_com}.}
        \label{tab:com_distance}
        \begin{tabular}{|l|c|c|}
            \hline
            \shortstack{\\Arm segment \\ (origin joint)}& \shortstack{Male \\ (\% segment length)}& \shortstack{Female \\ (\% segment length)}\\
            \hline
            Upper arm (shoulder)& 57.72& 57.54\\
            Forearm (elbow)& 45.74& 45.59\\
            Hand (wrist)& 79.00& 74.74\\
            \hline
        \end{tabular}
    \end{table}
    When holding an object of mass $m_{obj}$, the joint torques depend on the masses of arm segments between the joint and the object, as well as the object's weight. Let $m_{ua}$, $m_{fa}$, and $m_{h}$ denote masses of the upper arm, forearm and hand; $p_{ua}$ and $p_{fa}$, and $p_{h}$ denote the percentage distances of each segment's COM from its proximal joint origin; and $l_h$ denote the hand length. The reference torques at elbow ($\tau_e$) and shoulder ($\tau_s$) were calculated as follows:
    \begin{align}
            \tau_e &= m_{fa} \cdot g \cdot \frac{p_{fa}}{100} \cdot x \;+ \\  \nonumber
            &(m_{obj} + m_h) \cdot g \cdot \Big(l_{fa} + \frac{p_{h}}{100} \cdot l_h \cdot sin(\theta_e - \theta_s)\Big) \\
            \tau_s &= \tau_e + m_{ua} \cdot g \cdot \frac{p_{ua}}{100} \cdot y + (m_{obj} + m_h + m_{fa}) \cdot g \cdot l_{ua}
    \end{align}
    Furthermore, the shoulder torque $\tau_s$ was projected onto the sagittal and frontal planes to obtain
    \begin{math}
            \tau_{sf} = \tau_s \cdot cos(\theta_s)
    \end{math} and 
    \begin{math}
        \tau_{ss} = \tau_s \cdot sin(\theta_s)
    \end{math}.
    Based on these equations, torque profiles were derived for all experimental conditions described in Section~\ref{subsec:expt_procedure}. 

    \subsection{Feature Extraction Pipeline}
    Feature extraction is the process of transforming raw data into an optimal set of features that enhance the predictive performance of any machine learning model \cite{gallon2024comparison}. 
    This section details the steps followed to extract discriminative features from raw multi-channel EMG signals.

    \subsubsection{EMG Signal Preprocessing} \label{sec:preprocessing}
    Before extracting features, the EMG signals were processed to remove motion artifacts, baseline drifts, and other unwanted noise, thereby improving data quality for further analysis. 
    In this step, the raw EMG signals were filtered and smoothed following \cite{naeem2012emg}. Baseline offset was removed by subtracting the participant's rest signal from the raw data. To reduce noise and motion artifacts while preserving the main signal components, a zero-phase fourth-order Butterworth IIR band-pass filter was applied. As the EMG sampling frequency was \SI{500}{\si{Hz}}, the low-pass cutoff was set to \SI{225}{\si{Hz}} to avoid aliasing and the high-pass cutoff was set to \SI{15}{\si{Hz}}.

    After band-pass filtering, the EMG signals were further processed using a variance-based filter to capture amplitude fluctuations over time. This filter computed the running variance over a fixed-width window by incrementally updating the mean and variance as new samples entered and old samples were removed from a ring buffer. 
    This approach not only maximized computational efficiency but also smoothed out small fluctuations in EMG. 
    
    Next, the EMG signals were normalized using one of two strategies. In \emph{global normalization} (Condition A of Table~\ref{tab:results}), the signals from all weight and movement conditions were concatenated and normalized per channel using the maximum amplitude computed across the entire dataset. In \emph{condition-specific normalization} (Condition B of Table~\ref{tab:results}), maximum amplitude-based normalization was applied separately for each weight and movement condition. In both cases, the EMG signals were scaled to a range of 0 to 1. Finally, to further smooth the signal, a low-pass filter with a cutoff frequency of \SI{5}{\si{Hz}} was applied.

    \subsubsection{Neural Activation}
    When motor neurons in the spinal cord receive activation commands, they generate action potentials that induce electrical activity in the muscles \cite{Zayia2023MotorNeuron}. This activity ultimately produces muscle forces, but not instantaneously. A short time lag exists between the EMG-detected activity and the onset of muscle force production, known as the electromechanical delay ($d$), which varies across muscles and recording channels \cite{buchanan2004neuromusculoskeletal}. The resulting neural activation $p(t)$ was modeled using the following second-order difference equation as mentioned in \cite{buchanan2004neuromusculoskeletal, naeem2012emg}:
    \begin{equation}
        p(t) = \alpha \cdot e(t-d) + \beta_1 \cdot p(t-1) + \beta_2 \cdot p(t-2)
    \end{equation}
    where $e(t)$ is the rectified and smoothed EMG signal, and $\alpha$, $\beta_1$, and $\beta_2$ are the coefficients estimated using a parameter optimization function. To ensure that $p(t)$ remained bounded between 0 and 1, the coefficients must satisfy the following constraints:
    \begin{math}
        \alpha + \beta_1 + \beta_2 = 1, \quad
        \alpha \ge 0, \quad \beta_1 \ge 0, \quad \beta_2 \ge 0
    \end{math}.

    Subsequently, a sliding window technique was applied, using a segmentation window of \SI{100}{\milli\second} with 50\% overlap. Following \cite{igual2019myoelectric}, this represented the shortest effective window length, as shorter durations were shown to reduce system performance. The minimal length was chosen to facilitate future extension to online joint-torque estimation for rehabilitation scenarios, ensuring a balance between performance and computational efficiency. Feature extraction was performed independently on each window. It is important to note that for frequency-domain feature extraction, only the band-pass filtered signals were used to preserve the main frequency components and avoid distortions introduced by variance filtering.
    
    \subsubsection{Time-domain Features:}
    In addition to time points within each window, specific time-domain features were extracted, including root mean square (RMS), waveform length (WL), and slope sign change (SSC), as discussed in \cite{kok2024machine}. The mean absolute value (MAV) was initially considered but excluded because it was found to be highly correlated with RMS.

    \subsubsection{Frequency-domain Features}
    In addition to time-domain features, frequency band power (BP) was extracted to quantify the spectral power contained within specific frequency bands. BP was computed from the power spectral density (PSD) of the band-pass filtered EMG signals using the multitaper method \cite{teixeira2011epilab}. 
    
    \subsubsection{Time-frequency-domain Features}
    The limitations of using only time-domain or frequency-domain features are that for highly non-stationary signals, such as EMG, transient events and rapid changes in muscle activation are not well captured \cite{kok2024machine, cohen2002time}. Thus, time-frequency methods, such as the Morlet wavelet transform (MWT), were employed to quantify the evolution of spectral power within specific frequency bands over time, thereby enriching the feature set. In this work, the analysis frequencies were chosen from \SI{50}{\si{Hz}} to \SI{226}{\si{Hz}} in steps of 25, and the number of cycles $n_{cycles}$ was selected to satisfy the condition 
    \begin{math}
            n_{cycles} \cdot \frac{f_{sampling}}{f_{wavelet}} \le L_{window}
    \end{math},
    where $L_{window}$ is the length of the segmentation window in samples. 

    \subsubsection{Feature Postprocessing} The extracted features were split into training and test sets in a 90:10 ratio, with 15\% of the training data reserved for validation. History stacking was then performed within each split, ensuring that only windows from the same weight and movement type conditions were stacked together. This allowed the regressor to learn temporal dependencies, thereby improving the joint torque prediction. Furthermore, dimensionality reduction was performed using Principal Component Analysis (PCA).
    To prevent larger scales from dominating the PCA, the features were standardized to zero mean and unit variance. The reduced features were subsequently re-standardized to ensure consistent input scaling.
    
    To incorporate categorical features and improve model interpretability while preventing bias, one-hot encoding was applied \cite{datasciencedojo2024categorical}. This encoding technique represented each category as a separate binary column, wherein a value of 1 indicated presence and 0 indicated absence. Finally, the training feature set was shuffled just before being fed into the model.

    \subsubsection{Output Feature Extraction} Unlike the input features, the output requires minimal processing before it can be used as target values for supervised learning. However, certain steps were necessary to ensure consistency and stability in the training process. The torque values were first low-pass filtered with a cutoff frequency of \SI{5}{\si{Hz}} to remove high-frequency fluctuations and smooth the torque profiles. The same sliding window technique with \SI{100}{\milli\second} segmentation windows and 50\% overlap was then applied to the torque values of all three joints. The last sample of each window was selected as the feature. Finally, before being used as model targets, the three torque channels were scaled to $[-1,1]$ using a Min-Max scaler, to ensure that all outputs were on a consistent scale and facilitate stable training of the regressor.

    \subsection{Machine Learning Models} \label{sec:ml_models}
    To evaluate the performance of our feature extraction pipeline, we compared two distinct neural network architectures: a multilayer perceptron (MLP) network and a temporal convolutional network (TCN).
    \subsubsection{Multilayer Perceptron Network (MLP)} \label{sec:mlp}
\begin{figure}[!htbp]
      \centering
      \includegraphics[width=0.65\columnwidth]{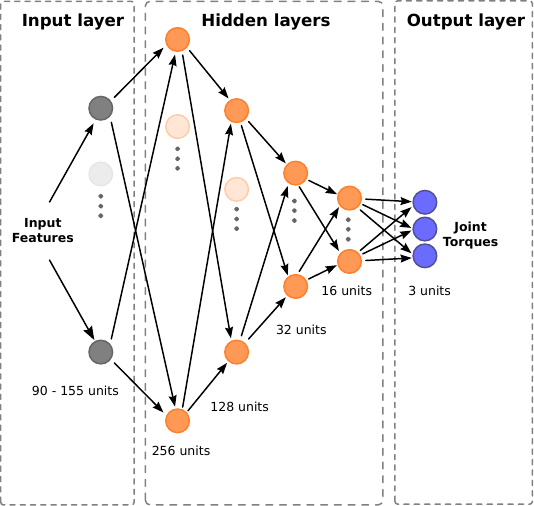}
      \caption{Structure of the MLP feedforward model.}
      \label{fig:mlp_arch}
    \end{figure}

    MLP is a supervised feedforward neural network consisting of an input layer, one or more hidden layers, and an output layer. The specific architecture employed in this work is shown in Figure~\ref{fig:mlp_arch}. It uses nonlinear activation functions in its hidden layers to model complex non-linear input-output mappings. To address the vanishing gradient problem, the ReLU activation function was chosen over sigmoid or tanh for the hidden layers, while the output layer used a tanh activation to match the $[-1,1]$ target scaling.
    To improve generalization and reduce overfitting, batch normalization (zero mean, unit variance), dropout (10\% random neuron deactivation), and $L_2$ regularization ($\lambda_2 = 0.001$) were applied.
    
    Furthermore, the targets were imbalanced: $\tau_e$ and $\tau_{sf}$ exhibited higher variance, and $\tau_{sf}$ and $\tau_{ss}$ had larger magnitudes. Thus, a weighted Huber loss function was used in place of MSE and the weights were defined as
    \begin{math}
        w_i = \frac{1}{\mathrm{Var}(\tau_i)+ \epsilon}
    \end{math} \cite{meyer2021alternative}.
    This combined approach avoided dominance of high-variance channels and ensured faster convergence and robust learning.
    
    \subsubsection{Temporal Convolutional Network} \label{sec:tcn}
    TCN is a convolutional architecture designed explicitly for sequence modeling tasks \cite{bai2018empirical}. It uses stacks of causal convolutions that ensure that the output at time $t$ depends only on inputs from time points $\leq t$. To efficiently expand the receptive field while keeping model complexity manageable, higher layers use dilated convolutions, complemented by residual connections to facilitate stable training. This architecture was chosen because \cite{bai2018empirical} shows that it outperformed recurrent architectures such as LSTMs or GRUs, even with minimal tuning. This makes TCN a viable solution for applications where models need to be retrained for each individual.
    
    The implemented TCN consisted of two residual stacks with a dilation factor of 1 and 2, respectively. Each stack contained one-dimensional convolutional layers with a kernel size of 3, 32 filters, ReLU activation, layer normalization, and a 10\% spatial dropout. A global average pooling layer followed the convolution blocks to condense the temporal feature maps into a fixed-length representation. The pooled output was then passed through a dense layer with 64 neurons and ReLU activation for feature refinement, before branching into three outputs predicting elbow, front-shoulder, and side-shoulder torques.
    The TCN was evaluated under two conditions: (i) using only preprocessed time points as input (Condition C of Table~\ref{tab:results}), as in \cite{zanghieri2021semg}, and (ii) using the complete proposed feature extraction pipeline (Condition B of Table~\ref{tab:results}).
    
    From Section~\ref{sec:mlp} and~\ref{sec:tcn}, it is evident that MLP is suited for general regression tasks with static features, whereas the TCN is specifically designed to capture temporal dependencies in sequential time-series data. Thus, comparing these two models allowed us to assess whether the proposed feature extraction pipeline could extract sufficient temporal information to enable even a feedforward model, such as MLP, to accurately predict joint torques from EMG signals. Furthermore, to ensure a fair comparison, signal processing parameters and core hyperparameters were kept constant across both models.

    \subsubsection{Filtering of Predicted Torques}
    The predicted torques were first processed using a median filter with a window size of 5, followed by a Savitsky-Golay filter \cite{schafer2011savitzky} with a window of size 21 and polynomial order 2. This two-step filtering process reduced noise while preventing over-smoothing of signal peaks. 

    \subsubsection{Evaluation Methodology}
    For evaluating the two models described in Section~\ref{sec:ml_models}, root mean squared error (RMSE), Pearson's correlation coefficient ($\rho$), and coefficient of determination ($R^2$) were used \cite{plevris2022investigation}. RMSE quantifies the average magnitude of prediction errors in the same unit as the target, making it easier to interpret and understand the errors. Pearson's $\rho$ ($[-1,1]$) measures the linear relationship between the target and predicted values, reflecting how well the model captures temporal dynamics; however, it does not account for the amplitude bias. In contrast, $R^2$ measures the variance explained by the predictions, capturing both temporal dynamics and amplitude accuracy. Thus, a lower RMSE and $\rho$, $R^2$ values closer to 1 indicate more accurate joint torque predictions.

    Furthermore, to ensure robust generalization, the test set included data from all weight and movement conditions, and five random seeds (1, 7, 25, 45, 70) were used for model initialization. We also enabled TensorFlow deterministic settings to fix weight initialization, shuffling, and dropout randomness across runs. The results are presented as mean and standard deviation across seeds, ensuring consistency in model performance and fair comparisons.

\section{RESULTS AND DISCUSSION} \label{sec:results}
This section presents the evaluation of the proposed feature extraction pipeline using 
the two machine learning models introduced in Section~\ref{sec:ml_models}. The evaluation was structured into three parts. 

\begin{table*}
\centering
\caption{Evaluation metrics for MLP and TCN under three input conditions. $^*$Condition A: \emph{global normalization} of concatenated EMG signals with the proposed feature extraction pipeline; Condition B: \emph{condition-specific normalization} with the proposed pipeline for feature extraction; Condition C: only preprocessed time points as input features.}
\label{tab:results}
\begin{tabular}{>{\centering\arraybackslash}p{6mm} p{13.5mm} >{\centering\arraybackslash}p{8mm} ccc ccc ccc}
\hline
\multirow{2}{*}{Model} & 
\multirow{2}{*}{\shortstack{Weights \\(kg)}} & 
\multirow{2}{*}{Condition$^*$} & 
\multicolumn{2}{c}{Elbow} & 
\multicolumn{2}{c}{Front} & 
\multicolumn{2}{c}{Side} \\
 & & & RMSE & $R^2$ & RMSE & $R^2$ & RMSE & $R^2$ \\
\hline
\multirow{4}{*}{MLP} 
 & 1.1, 1.85 & A & $1.119\pm0.238$ & $0.808\pm0.008$ & $2.215\pm0.090$ & $0.760\pm0.019$ & $2.046\pm0.103$ & $0.866\pm0.013$ \\
 & 1.1, 1.85 & B & $1.098\pm0.031$ & $0.802\pm0.011$ & $1.565\pm0.047$ & $0.810\pm0.011$ & $1.562\pm0.099$ & $0.877\pm0.015$  \\
 & 0, 1.1, 1.85 & A & $2.373\pm0.160$ & $0.235\pm0.102$ & $3.947\pm0.113$ & $0.275\pm0.041$ & $2.365\pm0.108$ & $0.843\pm0.014$  \\
 & 0, 1.1, 1.85 & B & $0.963\pm0.039$ & $0.853\pm0.012$ & $1.403\pm0.093$ & $0.833\pm0.023$ & $1.434\pm0.103$ & $0.875\pm0.018$  \\[2mm]
\multirow{3}{*}{TCN} 
 & 1.1, 1.85 & C & $1.136\pm0.035$ & $0.789\pm0.013$ & $1.757\pm0.052$ & $0.762\pm0.014$ & $1.657\pm0.186$ & $0.860\pm0.034$  \\
 & 0, 1.1, 1.85 & C & $0.964\pm0.028$ & $0.853\pm0.009$ & $1.563\pm0.038$ & $0.794\pm0.010$ & $1.494\pm0.043$ & $0.863\pm0.008$  \\
 & 0, 1.1, 1.85 & B & $0.933\pm0.049$ & $0.862\pm0.015 $& $1.459\pm0.055$ & $0.820\pm0.014$ & $1.314\pm0.083$ & $0.895\pm0.014$  \\
\hline
\end{tabular}
\end{table*}

In the first part, we investigated the effect of input normalization (\emph{global} versus \emph{condition-specific}) on MLP performance using \SI{1.1}{\si{kg}} and \SI{1.85}{\si{kg}} weights. Table~\ref{tab:results} (rows 1,2) reports the mean RMSE and $R^2$ metrics for \emph{global normalization} (Condition A) and \emph{condition-specific normalization} (Condition B), as described in Section~\ref{sec:preprocessing}. Apart from normalization, the feature extraction pipeline remained identical. Although performance differences were modest, \emph{condition-specific normalization} consistently outperformed \emph{global normalization}, likely due to the similar EMG amplitudes and torque profiles of the two heavier weights. To test this further, we added the \SI{0}{\si{kg}} condition, which has noticeably lower muscle activation and torque magnitudes, to the \emph{global normalization} strategy. This led to a substantial drop in mean $R^2$ for the elbow and front-shoulder joints (from 0.808 to 0.235, and from 0.760 to 0.275) as shown in Table~\ref{tab:results}, rows 1,3, while the side-shoulder performance remained largely unaffected. This can be attributed to the lower variance of shoulder abduction/adduction torques exhibited across weights and the more consistent activation patterns of the deltoid and pectoralis major muscles. Based on these results, \emph{condition-specific normalization} was adopted for all subsequent evaluations of MLP.

In the second part, we evaluated the TCN model using only normalized and smoothed time points as inputs (Condition C of Table~\ref{tab:results}), following \cite{zanghieri2021semg}, with the same \SI{1.1}{\si{kg}} and \SI{1.85}{\si{kg}} weights. As shown in Table~\ref{tab:results} (rows 1,2,5), TCN outperformed MLP with \emph{global normalization} (Condition A of Table~\ref{tab:results}) but slightly underperformed MLP with \emph{condition-specific normalization} (Condition B of Table~\ref{tab:results}), with RMSE values of 1.136, 1.757, and 1.657 for the elbow, front-shoulder, and side-shoulder joints, versus 1.098, 1.565, and 1.562. This behavior is expected as TCN is specifically designed to model temporal dynamics even from minimally processed time point inputs as discussed in Section~\ref{sec:tcn}.

\begin{figure*}[htbp]
    \centering
    \parbox{0.55\columnwidth}{
        \centering
        \subfloat[Elbow joint\label{fig:results_a}]{\includegraphics[width=\linewidth]{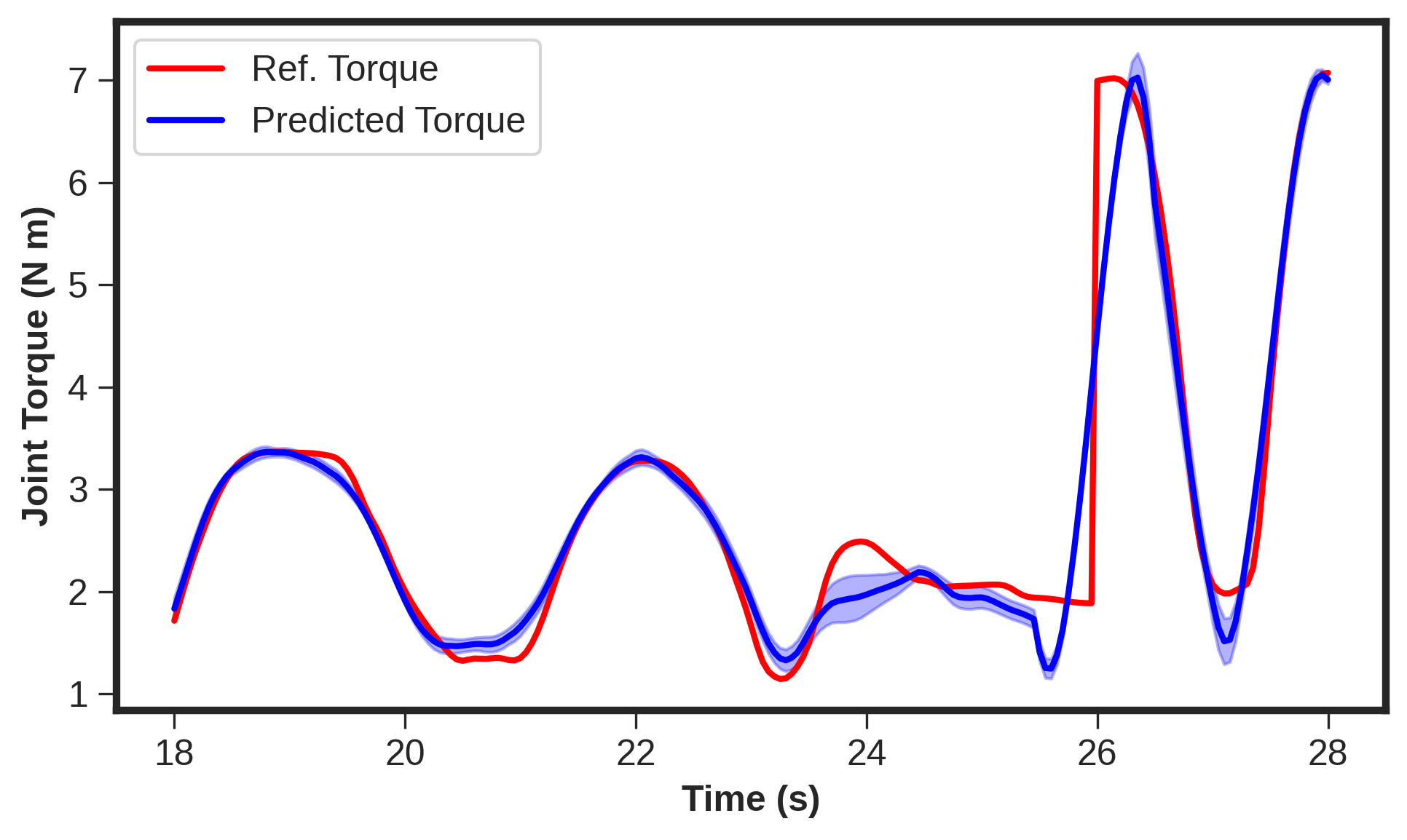}
        }
        }
    \parbox{0.55\columnwidth}{
        \centering
        \subfloat[Front-Shoulder joint\label{fig:results_b}]{\includegraphics[width=\linewidth]{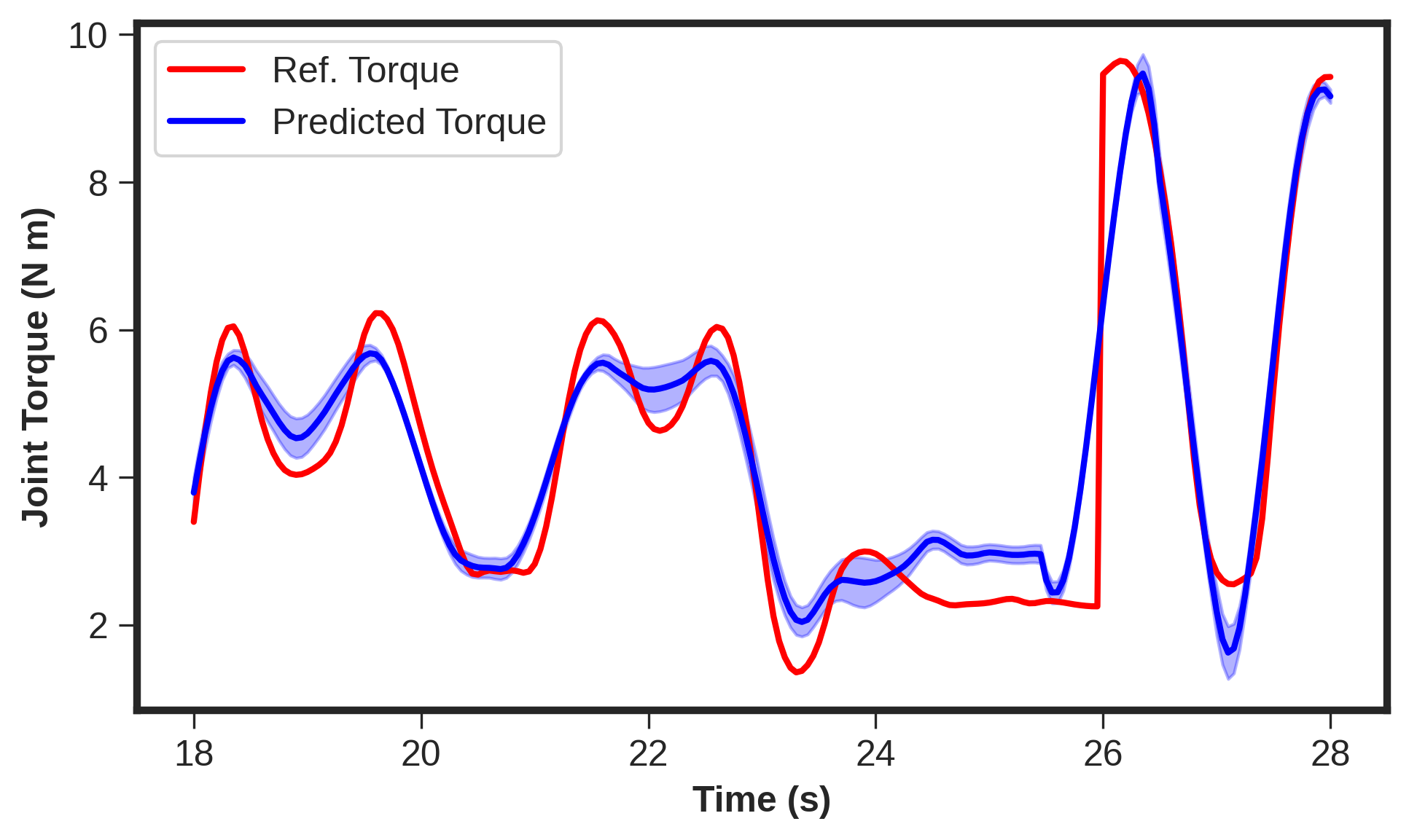}
        }
        }
    \parbox{0.55\columnwidth}{
        \centering
        \subfloat[Side-shoulder joint\label{fig:results_c}]{\includegraphics[width=\linewidth]{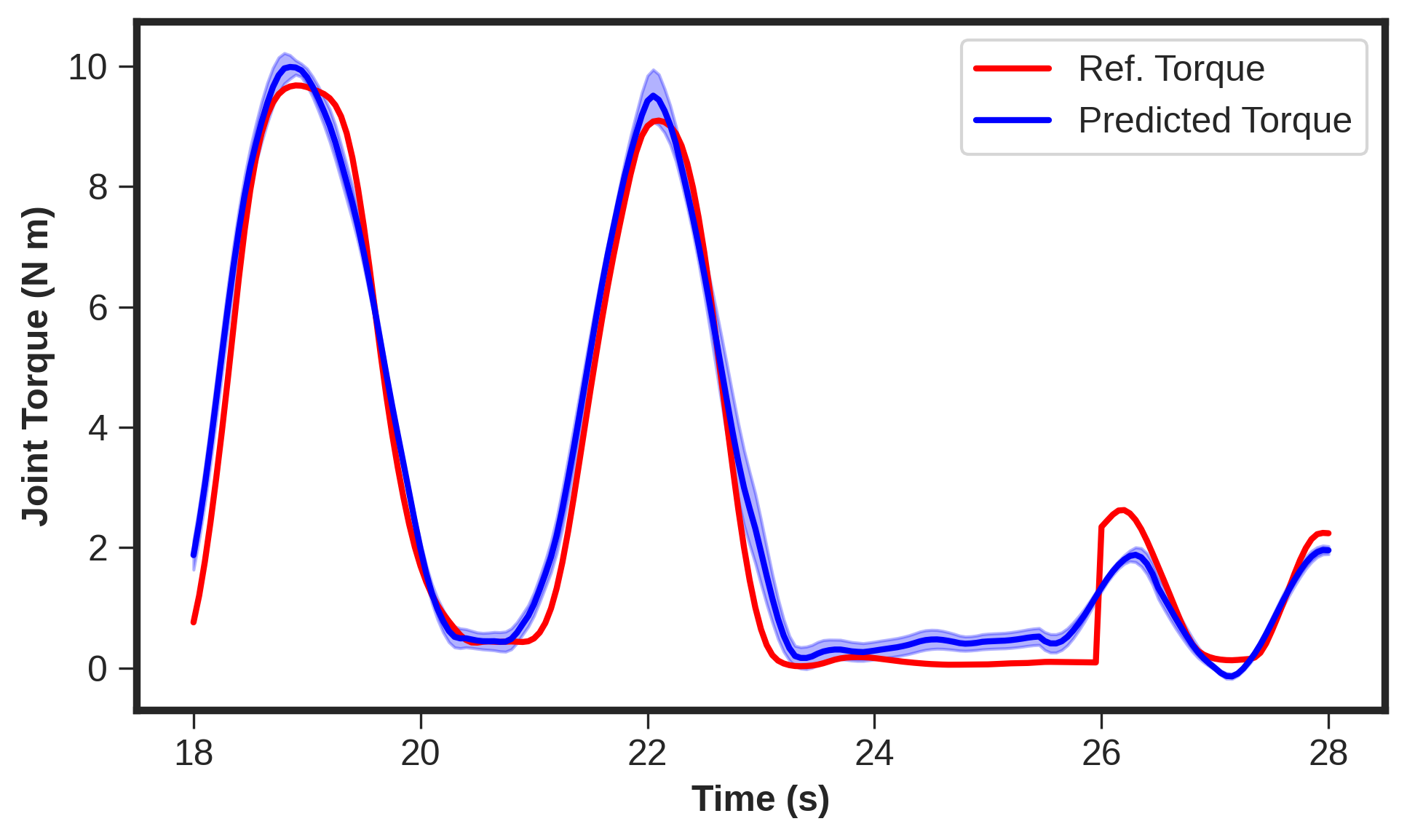}
        }
        }
    \hfill
    \parbox{0.55\columnwidth}{
        \centering
        \subfloat[Elbow joint\label{fig:results_d}]{\includegraphics[width=\linewidth]{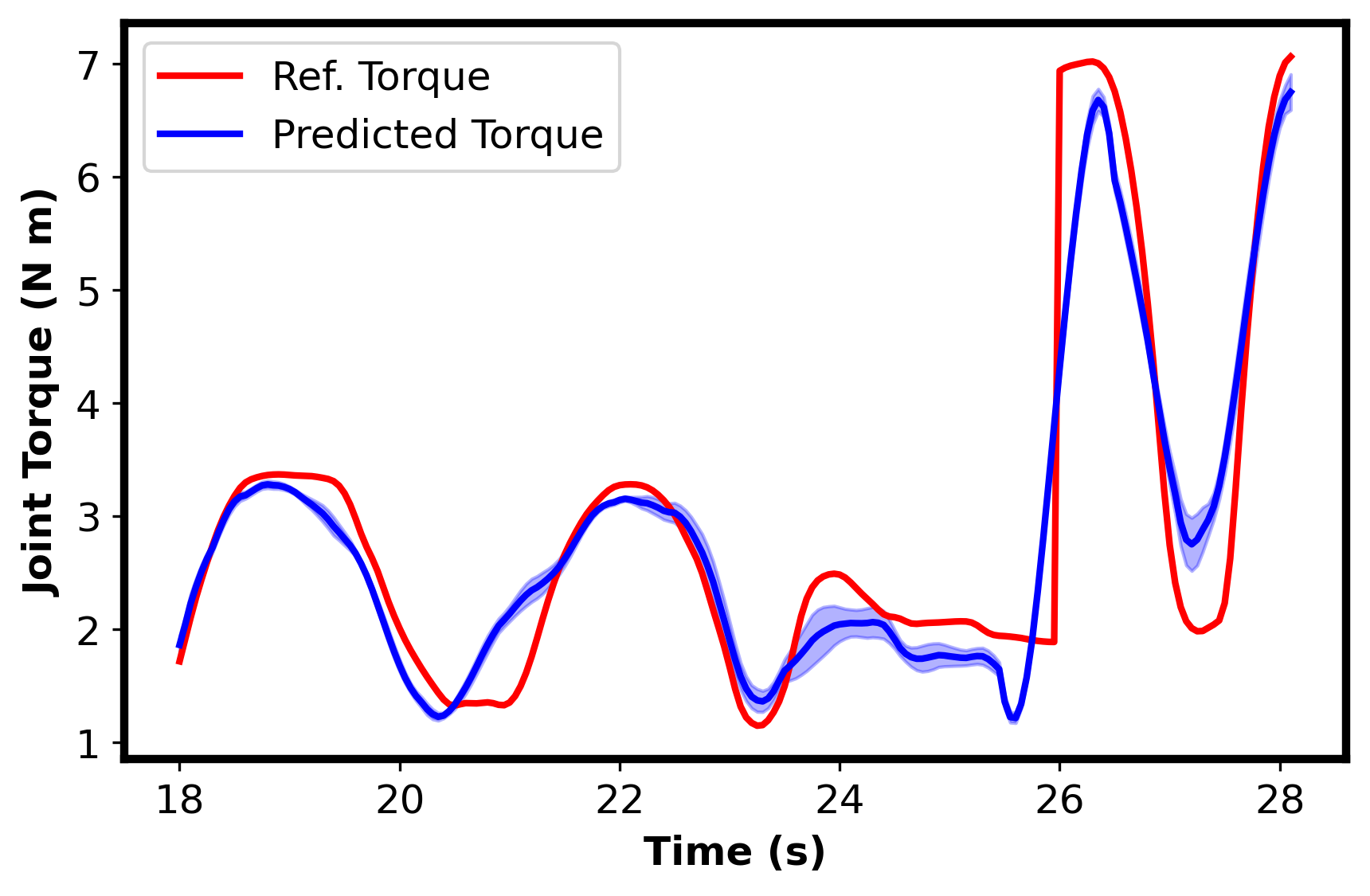}
        }
        }
    \parbox{0.55\columnwidth}{
        \centering
        \subfloat[Front-Shoulder joint\label{fig:results_e}]{\includegraphics[width=\linewidth]{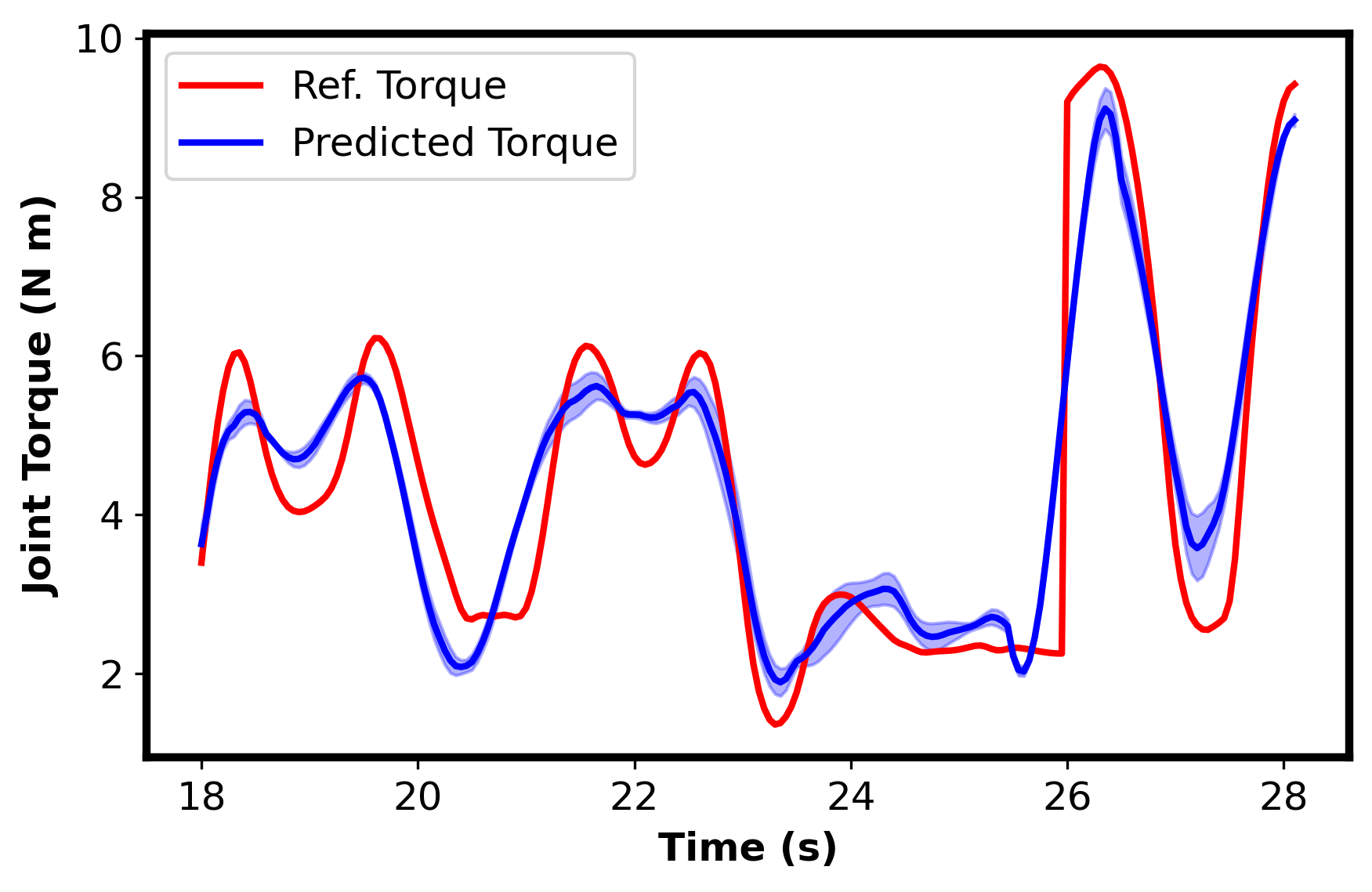}
        }
        }
    \parbox{0.55\columnwidth}{
        \centering
        \subfloat[Side-shoulder joint\label{fig:results_f}]{\includegraphics[width=\linewidth]{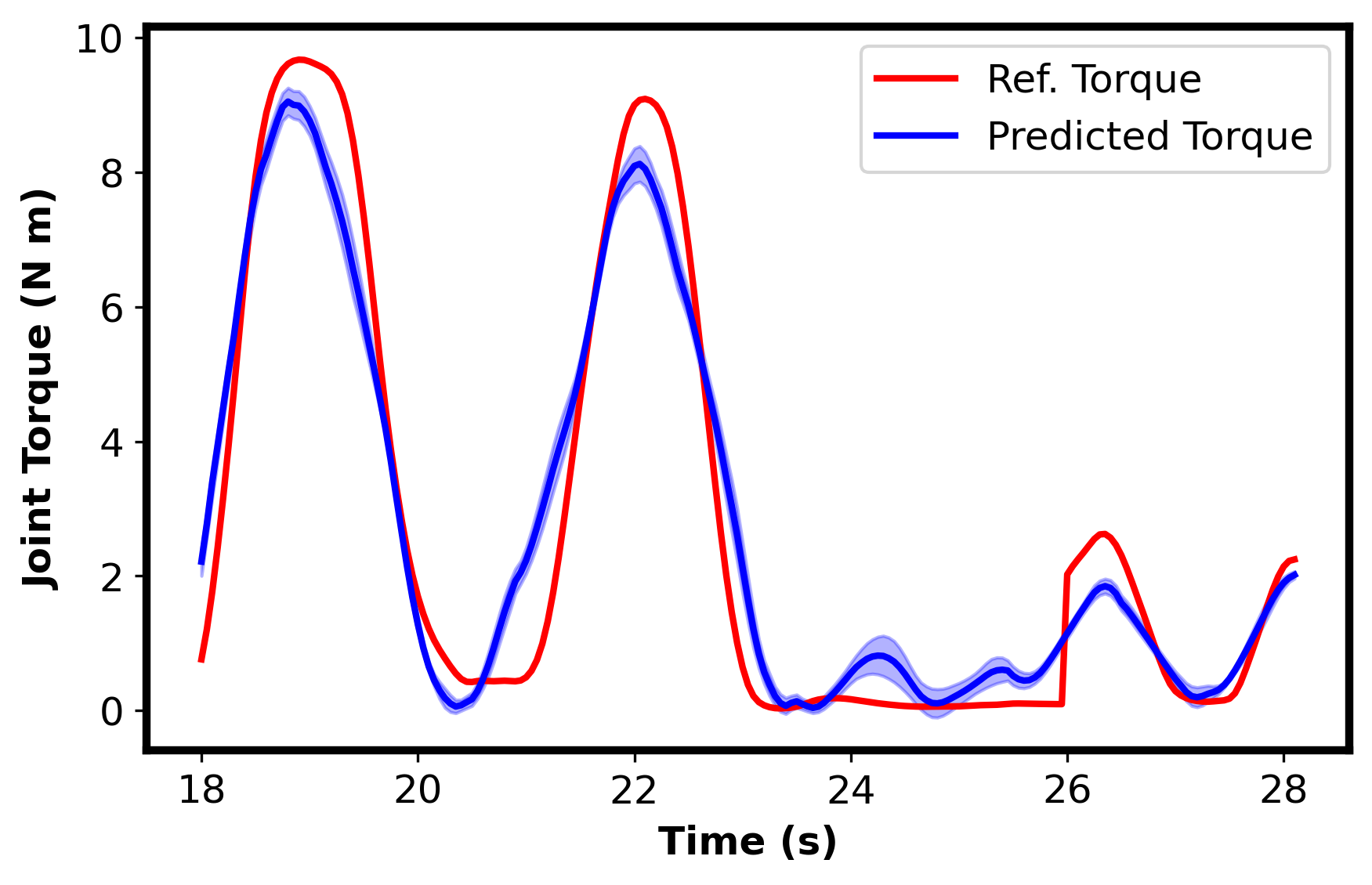}
        }
        }
    \caption{Band plot of predicted vs. reference joint torques for MLP and TCN models. Top row ((a)-(c)) shows MLP predictions, and bottom row ((d)-(f)) shows TCN predictions. The shaded band represents $\pm$ 1 standard deviation across five seeds around the mean torque.}
    \label{fig:results}
\end{figure*}

Finally, we compared MLP with the proposed feature extraction pipeline (Condition B of Table~\ref{tab:results}) against TCN using only preprocessed time points (Condition C of Table~\ref{tab:results}) on the complete dataset. As shown in Figure~\ref{fig:results}, MLP tracked the reference torques more closely than TCN across all joints. This is further supported by higher Pearson's $\rho$ values (mean $\pm$ standard deviation across five seeds: $0.926\pm0.006$, $0.915\pm0.013$, and $0.956\pm0.004$ for MLP versus $0.924\pm0.005$, $0.899\pm0.004$, and $0.933\pm0.002$ for TCN), indicating that MLP captured temporal dynamics more effectively than TCN. Similarly, the mean RMSE and $R^2$ values (Table~\ref{tab:results}, rows 4,6) confirm that MLP also better captured amplitude variations, particularly for the shoulder joints.

We further evaluated the impact of the proposed pipeline on TCN performance. The results show improvements, particularly for the shoulder joints, where the mean $R^2$ increased from 0.794 to 0.820 for the front-shoulder joint and from 0.863 to 0.895 for the side-shoulder joint (see Table~\ref{tab:results}, rows 6,7). Overall, these preliminary results indicate that the proposed pipeline enabled even a simple MLP network, despite lacking explicit temporal modeling, to achieve performance comparable to or better than TCN, particularly for the front-shoulder joint. 

\section{CONCLUSION} \label{sec:conclusion}
Accurate joint torque estimation is essential for safe robot-assisted rehabilitation. We proposed a feature extraction pipeline to enhance EMG-based joint torque prediction and evaluated it with MLP and TCN models. 
The preliminary results demonstrated that 
our feature extraction pipeline is effective in combination with lightweight neural networks, which makes it suitable for conditions where only limited training data is available.
Future work will focus on improved reference torque calculation using inverse dynamics, online implementation, and systematic user studies.

\bibliographystyle{IEEEtran}
\bibliography{references}

\end{document}